\documentclass[runningheads]{llncs}
\usepackage{graphicx}
\usepackage{amsmath,graphicx}
\usepackage{times}
\usepackage{latexsym}
\usepackage{times}
\usepackage{latexsym}
\usepackage{caption}
\usepackage{subcaption}
\usepackage{comment}
\usepackage{times}
\usepackage{soul}
\usepackage{booktabs}
\usepackage{url}
\usepackage[utf8]{inputenc}
\usepackage{graphicx}
\usepackage{amsmath}
\usepackage{amsmath,graphicx}
\usepackage{times}
\usepackage{latexsym}
\usepackage{times}
\usepackage{latexsym}
\usepackage{comment}
\usepackage{times}
\usepackage{soul}
\usepackage{booktabs}
\usepackage{url}
\usepackage[utf8]{inputenc}
\usepackage{graphicx}
\usepackage{amsmath}
\usepackage{amssymb}
\usepackage{array}
\usepackage{booktabs}
\usepackage{algorithm}
\usepackage{algorithmic}
\usepackage{multirow}
\usepackage{multicol}
\usepackage{caption}
\usepackage{xspace}
\usepackage{amssymb}
\usepackage{booktabs}
\usepackage{algorithm}
\usepackage{algorithmic}
\usepackage{multirow}
\usepackage{multicol}
\usepackage{caption}
\usepackage{xspace}
\usepackage{amsmath}

\DeclareMathOperator*{\argmin}{arg\,min}
\usepackage{url}
\usepackage{url}
\newcommand\blfootnote[1]{%
  \begingroup
  \renewcommand\thefootnote{}\footnote{#1}%
  \addtocounter{footnote}{-1}%
  \endgroup
}
\usepackage[symbol]{footmisc}

\renewcommand{\thefootnote}{\fnsymbol{footnote}}

\usepackage{fontawesome}

\makeatletter
\newcommand{\printfnsymbol}[1]{%
  \textsuperscript{\@fnsymbol{#1}}%
}
\makeatother
\begin{document}
\title{A SentiWordNet Strategy for Curriculum Learning in Sentiment Analysis}%
\author{Vijjini Anvesh Rao\printfnsymbol{1} \and Kaveri Anuranjana\printfnsymbol{1} \and Radhika Mamidi}
\institute{Language Technologies Research Center \\ Kohli Center on Intelligent Systems\\ International Institute of Information Technology, Hyderabad, India
          \email{\{vijjinianvesh.rao,kaveri.anuranjana\}@research.iiit.ac.in \\ radhika.mamidi@iiit.ac.in}}

%
%
%
%
\maketitle              
\begin{abstract}
Curriculum Learning (CL) is the idea that learning on a training set sequenced or ordered in a manner where samples range from easy to difficult, results in an increment in performance over otherwise random ordering. The idea parallels cognitive science's theory of how human brains learn, and that learning a difficult task can be made easier by phrasing it as a sequence of easy to difficult tasks. This idea has gained a lot of traction in machine learning and image processing for a while and recently in Natural Language Processing (NLP). In this paper, we apply the ideas of curriculum learning, driven by SentiWordNet in a sentiment analysis setting. In this setting, given a text segment, our aim is to extract its sentiment or polarity. SentiWordNet is a lexical resource with sentiment polarity annotations. By comparing performance with other curriculum strategies and with no curriculum, the effectiveness of the proposed strategy is presented. Convolutional, Recurrence, and Attention-based architectures are employed to assess this improvement. The models are evaluated on a standard sentiment dataset, Stanford Sentiment Treebank.
\blfootnote{\printfnsymbol{1}These authors have contributed equally to this work}

\end{abstract}

\section{Introduction}
Researchers from Cognitive Science have established a long time ago that humans learn better and more effectively in an incrementive learning setting \cite{skinner1958reinforcement,krueger2009flexible}. Tasks like playing a piano or solving an equation, are learnt by humans in a strategy where they are first provided easier variants of the main challenge, followed by gradual variation in difficulty. This idea of incremental human learning has been studied for machines as well, specifically in machine learning. Curriculum Learning (CL) as defined by \cite{bengio2009curriculum} introduce and formulate this concept from Cognitive Science to a machine learning setting. They observe that on shape recognition problem (rectangle, ellipse or triangle), training the model first on a synthetically created dataset with less variability in shape, generalizes faster as compared to directly training on the target dataset. Furthermore, other experiments by \cite{bengio2009curriculum} demonstrate performance improvements on a perceptron classifier when incremental learning is done based on the margin in support vector machines (SVM) and a language model task where  growth in vocabulary size was chosen as the curriculum strategy.
These examples indicate that while curriculum learning is effective, the choice of the curriculum strategy, the basis for ordering of samples is not clear cut and often task specific. Furthermore some recent works like \cite{mccann2018natural} have suggested that anti curriculum strategies perform better, raising more doubts over choice of strategy. In recent years Self-Paced Learning (SPL) \cite{kumar2010self,jiang2015self} has been proposed as a reformulation of curriculum learning by modeling the curriculum strategy and the main task in a single optimization problem. 
\\
Sentiment Analysis (SA) is a major challenge in Natural Language Processing. It involves classifying text segments into two or more polarities or sentiments. Prior to the success of Deep Learning (DL), text classification was dealt using lexicon based features. However sentiment level information is realized at more levels than just lexicon or word based. For a model to realize a negative sentiment for ``not good'', it has to incorporate sequential information as well. Since the advent of DL, the field has been revolutionized. Long Short Term Memory \cite{hochreiter1997long,madasu2019sequential} (LSTM), Convolutional Neural Networks (CNN) \cite{kim2014convolutional,madasu2019gated,madasu2019effectiveness} and Attention based architectures \cite{yang2016hierarchical,madasu2020position} have achieved state-of-art results in text classification and continue to be strong baselines for text classification and by extension, Sentiment Analysis. Sentiment Analysis, further aids other domains of NLP such as Opinion mining and Emoji Prediction\cite{choudhary2018twitter}.
\\
Curriculum Learning has been explored in the domain of Computer Vision (CV) extensively \cite{lee2011learning,jiang2014easy,imagecurr} and has gained traction in Natural Language Processing (NLP) in tasks like Question Answering \cite{sachan2016easy,sachan2018self}, Natural Answer Generation \cite{liu2018curriculum} and Domain Adaptation\cite{madasu2020sequential}. In Sentiment Analysis, \cite{cirik2016visualizing} propose a strategy derived from sentence length, where smaller sentences are considered easier and are provided first. \cite{han2017tree} provide a tree-structured curriculum based on semantic similarity between new samples and samples already trained on. \cite{bayesian2016learning} suggest a curriculum based on hand crafted semantic, linguistic, syntactic features for word representation learning. However, these CL strategies pose the easiness or difficulty of a sample irrespective of sentiment. While their strategies are for sentiment analysis, they do not utilize sentiment level information directly in building the order of samples. Utilizing SentiWordNet, we can build strategies that are derived from sentiment level knowledge.

SentiWordNet \cite{esuli2006sentiwordnet,baccianella2010sentiwordnet} is a highly popular word level sentiment annotation resource. It has been used in sentiment analysis and related fields such as opinion mining and emotion recognition. This resource was first created for English and due to its success it has been extended to many other languages as well \cite{das2010sentiwordnet,das2010towards,parupalli2018bcsat,parupalli2018towards}. This lexical resource assigns positivity, negativity and derived from the two, an objectivity score to each WordNet synset \cite{miller1995wordnet}. The contributions of the paper can be summarized as follow:

\begin{itemize}
    \item We propose a new curriculum strategy for sentiment analysis (SA) from SentiWordNet annotations.
    \item Existing curriculum strategies for sentiment analysis rank samples with a difficulty score impertinent to the task of SA.  Proposed strategy ranks samples based on how difficult assigning them a sentiment is. Our results show such a strategy's effectiveness over previous work.
\end{itemize}

While Curriculum Learning as defined by \cite{bengio2009curriculum} is not constrained by a strict description, later related works \cite{cirik2016visualizing,han2017tree,spitkovsky2010baby} make distinction between Baby Steps and One-Pass methods for enabling curriculum learning. In our experiments we observe better performance with Baby Steps over One Pass hence we use the same for proposed SentiWordNet driven strategy. \cite{cirik2016visualizing} also show that Baby Steps outperforms One Pass.
For every sentence $s_{i}$ $\in$ $D$, its sentiment is described as $y_{i}$ $\in$ $\{0,1,2,3,4\}$\footnote{Our dataset has 5 labels.}, where $i$ $\in$ $\{1,2,..,n\}$ for $n$ data points in $D$. For a model $f_{w}$, its prediction based on $s_{i}$ will be $f_{w}(s_{i})$. Loss $L$ is defined on the model prediction and actual output as $L(y_{i},f_{w}(s_{i}))$ and the net cost for the dataset is defined as $C(D,f_{w})$ as $\sum_{\forall i} \frac{1}{n}L(y_{i},f_{w}(s_{i}))$. Then the task is modelled by
\begin{equation}
    \min_{w} C(D,f_{w}) + g(w)
\end{equation}
Where $g(w)$ can be a regularizer. In this setting, Curriculum Learning is defined by a Curriculum Strategy $S(s_{i})$. $S$ defines an ``easiness'' quotient of sample $s_{i}$. If the model is currently trained on $D\prime$ $\subset$ $D$. Then sample $s_{j}$ $\in$ $D-D\prime$ is chosen based on $S$ as: 
\begin{equation}
    s_{j} = \argmin_{s_{i}\in D-D\prime} S(s_{i})
\end{equation}
Sample $s_{j}$ is then added to the new training set or $D\prime$ $=$ $D\prime$ $+$ $s_{j}$\footnote{Adding one sample at a time can be a very slow process, hence we add in batches. For our experiments, we take a batch size of $bs$ samples with lowest $S$ to add at once.} and the process continues until training is done on all the sentences in $D$. The process starts with first training on a small subset of $D$, which have least $S$ score. In this way incremental learning is done in Baby Steps.
\section{Experiments}
\subsection{Dataset}
Following previous works in curriculum driven sentiment analysis \cite{cirik2016visualizing,han2017tree,bayesian2016learning} We use the Stanford Sentiment Treebank (SST) dataset \cite{socher2013recursive}\footnote{https://nlp.stanford.edu/sentiment/}. Unlike most sentiment analysis datasets with binary labels, SST is for a 5-class text classification which consists of 8544/1101/2210 samples in train, development and test set respectively. We use this standard split with reported results averaged over 10 turns. 

\begin{table*}[t]
  \centering
  \setlength{\tabcolsep}{4pt}
  \renewcommand{\arraystretch}{1.25}
  \begin{tabular}{{c c c c}}
    \hline
     Model&  & Curriculum Strategies \\
    
    & SentiWordNet & Sentence Length & No Curriculum\\
    \hline
     Kim CNN  &\textbf{41.55} & 40.81 & 40.59  \\
    \hline
   LSTM & \textbf{44.54} & 43.89  & 41.71\\
    \hline
      LSTM+Attention  & \textbf{45.27} & 42.98 & 41.66  \\
    \hline
     \hline
  \end{tabular}
  \caption{Accuracy scores in percentage of all models on different strategies}
  \label{tab:Accuracy}
\end{table*}

\subsection{Architectures}
We test our curriculum strategies on popular recurrent and convolutional architectures used for text classification. It is imperative to note that curriculum strategies are independent of architectures, they only decide the ordering of samples for training. The training itself could be done with any algorithm.
\subsubsection*{Kim CNN}
This baseline is based on the deep CNN architecture \cite{kim2014convolutional} highly popular for text classification.
\subsubsection*{LSTM}
We employ Long Short Term Memory Network (LSTM) \cite{hochreiter1997long} for text classification. Softmax activation is applied on the final timestep of the LSTM to get final output probability distributions. Previous approach \cite{cirik2016visualizing} uses LSTM for this task as well with sentence length as curriculum strategy.
\subsubsection*{LSTM + Attention}
In this architecture we employ attention mechanism described in \cite{bahdanau2014neural} over the LSTM outputs to get a single context vector, on which softmax is applied.
In this baseline, attention mechanism is applied on the top of LSTM outputs across different timesteps. Attention mechanism focuses on most important parts of the sentence that contribute most to the sentiment, especially like sentiment words.
\subsection{Implementation Details}
We used GloVe pretrained word vectors\footnote{https://nlp.stanford.edu/data/glove.840B.300d.zip} for input embeddings on all architectures. The size of the word embeddings in this model is 300. A maximum sentence length of 50 is considered for all architectures. Number of filters taken in the CNN model is 50 with filter size as 3, 4, 5. We take number of units in the LSTM to be 168, following previous work \cite{tai2015improved} for empirical setup. For the LSTM + Attention model, we take number of units in the attention sub network to be 10. Categorical crossentropy as the loss function and Adam with learning rate of 0.01 as optimizer is used. The batch size $bs$ defined in curriculum learning framework is $900$ for Sentence Length Strategy in LSTM and LSTM+Attention, and $750$ for CNN. For SentiWordNet strategy, it is $1100$ for LSTM and LSTM+Attention, and $1400$ for CNN.

\subsection{Curriculum Strategies}
In this section we present the proposed SentiWordNet driven strategy followed by a common strategy based on sentence length.
\begin{figure}
\centering
  \includegraphics[width=10cm]{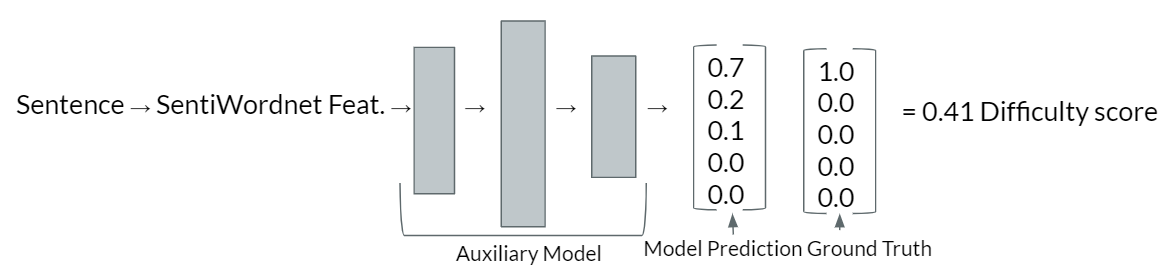}
  \caption{SentiWordNet based difficulty score illustrated.}
\end{figure}
\label{fig:strat}
\section{Problem Setting for Curriculum Learning}

\subsubsection{SentiWordNet driven Strategy}
 We first train an auxiliary feed forward model $Aux$ for sentiment analysis on the same dataset utilizing only SentiWordNet features. This allows us to find out which training samples are actually difficult. Following are the features we use for the auxiliary model:
\begin{itemize}
    \item Sentence Length $l$: For a given sentence, this feature is just the number of words after tokenization.
    \item Net Positivity Score $P$: For a given sentence, this is the sum of all positivity scores of individual words.
    \item Net Negativity Score $N$: For a given sentence, this is the sum of all negativity scores of individual words.
    \item Net Objectivity Score $O$: For a given sentence, this is the sum of all objectivity scores of individual words.\footnote{Note that for an individual word, The \textit{Objectivity score} is just \textit{1 - Negativity Score  - Positivity Score}}. This feature is meant to show how difficult it is to tell the sentiment of a sentence.
    \item Abs. Difference Score: This score is the absolute difference between Net Positivity and Net Negativity Scores or $AD = |P - N|$. This feature is meant to reflect overall sentiment of the sentence.
    \item Scaled Positivity: Since the Net Positivity may increase with number of words, we also provide the scaled down version of the feature or $\frac{P}{l}$.
    \item Scaled Negativity: For the same reason as above, we also provide $\frac{N}{l}$.
    \item Scaled Objectivity: Objectivity scaled down with sentence length or $\frac{O}{l}$.
    \item Scaled Abs Difference: Abs. Difference $D$ scaled down with sentence length or $\frac{AD}{l}$.
\end{itemize}
Since all the features lie in very different ranges, before passing for training to the architectures they are normalized between $-1$ and $1$ first with mean $0$. Also important to note is that, the SentiWordNet scores are for a synset and not for a word. In essence, a word may have multiple scores. In such cases, positivity, negativity and objectivity scores are averaged for that word. We use a simple feed forward network to train this auxiliary model with final layer as a softmax layer\footnote{The number of layer units are as follows: $[8,100,50,5]$.}. We get an accuracy of just $25.34$ on this model, significantly lesser than performances we see by LSTM and CNN in No Curriculum setting as seen in Table \ref{tab:Accuracy}. But the performance doesn't actually matter. From this model, we learn what samples are the most difficult to classify and what are the easiest. For all 8544 training samples of $D$, we define the curriculum score as follows:
\begin{equation}
S(s_i) = \sum_j^c(Aux(s_i)^j - y_i^j)^2
\end{equation}
where $Aux(s_i)^j$ is the prediction of auxiliary model $Aux$ on sentence $s_i$, $j$ is the iterator over the number of classes $c=5$. In essence, we find the mean squared error between the prediction and the sentence's true labels. If $S(s_i)$ is high, it implies the sentence is hard to classify and if less, then the sentence is easy. Because the features were trained on an auxiliary model from just SentiWordNet features, we get an easiness-difficulty score purely from the perspective of sentiment analysis. Figure \ref{fig:strat} shows an example difficulty score calculation with this strategy.
\subsubsection{Sentence Length}
This simple strategy tells that, architectures especially like LSTM find it difficult to classify sentences which are longer in length. And hence, longer sentences are difficult and should be ordered later. Conversely shorter sentence lengths are easier and should be trained first. This strategy is very common and has not only been used in sentiment analysis \cite{cirik2016visualizing}\footnote{\cite{cirik2016visualizing} have done CL on SST as well, however our numbers do not match because they use the phrase dataset which is much larger.} but also in dependency parsing \cite{spitkovsky2010baby}. Which is why it becomes a strong baseline especially to evaluate the importance of SentiWordNet driven strategy.

\section{Results}
We report our results in Table \ref{tab:Accuracy}. As evident from the table, we see that proposed SentiWordNet based strategy beats Sentence Length driven and No Curriculum always. However, the difference 
between them is quite less for the CNN model. This must be because, CNN for this dataset is worse of all other models without curriculum, this architecture finds it difficult to properly classify, let alone fully exploit curriculum strategies for better generalization. Furthermore, another reason behind effectiveness of Sentence Length strategy for LSTM and LSTM+Attention is that, considering the LSTM's structure which observes one word at a time, its only natural that longer sequences will be hard to remember, hence Sentence Length ordering acts as a good curriculum basis. This idea has also been referenced by previous works such as \cite{cirik2016visualizing}. Since Attention mechanism observes all time steps of the LSTM, the difficulty in longer sentence lengths diminishes and hence the improvement in performance with Sentence Length strategy is lesser as compared to LSTM. Sentence Length driven Strategy, while performing better in LSTM and LSTM+Attention model, is still less than SentiWordNet, this is because sentence length strategy defines difficulty and easiness in a more global setting, not specific to sentiment analysis. However, with SentiWordNet we define a strategy which characterizes the core of curriculum learning in sentiment analysis, namely the strategy for ranking samples based solely on how difficult or easy it is to classify the sample into predefined sentiment categories.
\section{Conclusion}
In this paper, we define a SentiWordNet driven strategy for curriculum learning on sentiment analysis task. The proposed approach's performance is evident on multiple architectures, namely recurrent, convolution and attention based proving the robustness of the strategy. This approach also shows the effectiveness of simple lexicon based annotations such as SentiWordNet and how they can be used to further sentiment analysis. Future works could include strategies that consecutively enrich SentiWordNet as well and also those that can refine the resource by pointing out anomalies in the annotation.
\bibliographystyle{splncs04}
\bibliography{nldb}
\end{document}